# Comparing SVM and Naïve Bayes Classifiers for Text Categorization with Wikitology as knowledge enrichment

Sundus Hassan
Computer Science Department
NUCES-FAST, Karachi Campus
K093059@nu.edu.pk

Muhammad Rafi
Computer Science Department
NUCES-FAST, Karachi Campus
muhammad.rafi@nu.edu.pk

Muhammad Shahid Shaikh
Electrical Engineering Department
NUCES-FAST, Karachi Campus
shahid.shaikh@nu.edu.pk

*Abstract*—The activity of labeling of documents according to their content is known as text categorization. Many experiments have been carried out to enhance text categorization by adding background knowledge to the document using knowledge repositories like Word Net, Open Project Directory (OPD), Wikipedia and Wikitology. In our previous work, we have carried out intensive experiments by extracting knowledge from Wikitology and evaluating the experiment on Support Vector Machine with 10- fold cross-validations. The results clearly indicate Wikitology is far better than other knowledge bases. In this paper we are comparing Support Vector Machine (SVM) and Naïve Bayes (NB) classifiers under text enrichment through Wikitology. We validated results with 10-fold cross validation and shown that NB gives an improvement of +28.78%, on the other hand SVM gives an improvement of +6.36% when compared with baseline results. Naïve Bayes classifier is better choice when external enriching is used through any external knowledge base.

*Keywords- Text Categorization, Machine Learning, Wikitology, Support Vector Machine, 20 News Group. Knowledge base, Naïve Bay*

## I. INTRODUCTION

The action of assigning one or more pre-defined categories to a document on the basis of its content is defined as text categorization. It has 2 basic types: Single label and Multi-label. It is formally defined as, assume there are set of Documents D= {d1, d2 …….dn}, where 'n' is very large number. Now by using any text categorization algorithm, the document has to be semantically categorized in pre-defined set of classes or categories C= {C1, C2…..Ck}, where each Ci is class. It is possible that documents categorized in 2 different categories such as category Ci and Cj, semantically may share largely some similar sense or they can be largely dissimilar also. Knowledge Engineering and Machine Learning are 2 main approaches to develop text categorization techniques. In this paper we are comparing results of Support Vector Machine (SVM) and Naïve Bayes techniques. Different Experimental results prove that SVM performs better than NB in general classification tasks. Many experiments have been carried out by researchers to enhance text categorization. Improvement is shown by adding semantic background to document. This method is used by researchers using different knowledge repositories like WordNet, OPD, Wikipedia and Wikitology etc.

Word Net is a lexical database of English language. Elberrichi, Rahmoun, Bentaalah [4] uses Word Net; Gabrilovich, Markovitch [5] uses OPD, Gabrilovich[6][8] and Pu Wang[7] uses Wikipedia knowledge to improve text categorization by retrieving knowledge from these resources. Hotho[9] incorporates Word Net knowledge to text documents to improve text clustering by reducing the variance of text semantics among documents under one category. Wikitology is a knowledge base where information is extracted from Wikipedia and kept in structured and unstructured form as information retrieval indices. In our previous work [12], we retrieved knowledge from Wikitology and added to document using different text document representation and enrichment techniques. We carried out intensive experiments by using combination of different text document representation and enrichment techniques. Experimental results proved that there was a better text categorization and clustering by enriching documents with the knowledge of Word Net, OPD, Wikipedia and Wikitology as compared to their defined baseline on diverse datasets. But the best text categorization was done by integrating knowledge from Wikitology.

In [12] an improvement of +6.36% (as compared to baseline) was shown by using machine learning technique Support Vector Machine (SVM). In this paper we have directed our research by using another machine learning technique Naïve Bayes. It has been shown that as Naïve Bayes shown an improvement of +28.78% (as compared to baseline).

The organization of our paper is as follows: Section 2 discusses the related work carried out in this domain. Section 3 illustrates empirical evaluation of experiments carried out on a dataset of 20 News Group.

## II. RELATED WORK

Many researchers shown text enriching method facilitates text categorization by adding semantic background knowledge from knowledge bases like Word Net, Wikipedia, Open Project Directory (OPD) and Wikitology.

Elberrichi, Rahmoun, Bentaalah [4] proposed a model in which for all the terms the general concepts are extracted from Word Net. Proposed model was evaluated on Reuters-21578 and 20 News Group dataset. Improvement from 0.649 to 0.714 and 0.667 to 0.719 is shown on Reuters-21578 and 20 News Group dataset respectively. Improvement in text clustering was also shown by Hotho, Staab, Stumme [9], incorporating knowledge from Word Net. They assimilate background information to text document to reduce the variance in semantics of a document under one category. Evaluation was done on Reuters-21578 using purity and inverse purity measures. Gabrilovich, Markovitch [5] suggested a model in which feature generator is built by mapping terms of document to Open Project Directory (OPD) concepts and features are generated by contextual analysis of the document. Model is evaluated on datasets of Reuters-21578, RCV1, 20 News Group and Movies. Evaluated results shown improvement on all the datasets.

Gabrilovich, Markovitch [6] also utilizes Wikipedia knowledge and put forward the model in which feature generation is done through multi-resolution approach. In this approach basically documents are matched with most relevant documents without considering thesaurus of the article. Proposed model is evaluated on diverse datasets (Reuters-21578, RCV1, OHSUMED, 20 News Group and Movies) show improvement. Wang, Hu, Zeng [7] focused on Wikipedia thesaurus like synonymy, polysemy, hyponymy and associative relations. Text document is enriched by integrating Wikipedia thesaurus and it is evaluated on Reuters-21578, OHSUMED and 20 News Group dataset. This approach shows more improvement then Gabrilovich, Markovitch [6] proposed model. Gabrilovich, Markovitch [8] proposed a novel method of semantic relatedness is calculated using Wikipedia-based explicit semantic analysis, in which concepts obtained from Wikipedia are represented in high-dimensional space. Improvement is shown in computing word or text relatedness. Razvan Bunescu and Marius Pasca [10] used Wikipedia information for resolving Named Entity Disambiguation. For this intention they exploit redirect pages, disambiguation pages, categories and hypelinks. They developed novel approaches from Wikipedia context article similarity, taxonomy and identifying entities which are out from Wikipedia. The best experimental results were obtained from Wikipedia taxonomy.

In this paper, we are extending the research present by us in [12]. In [12] we enrich document by adding knowledge from Wikitology. Wikitology is a hybrid knowledge base of structured and unstructured information extracted from Wikipedia. Knowledge in Wikitology is represented in different forms like an IR index, graphs (category links, page links and entity links), relational database and a triple store. We also present 4 document representation and 5 enrichment techniques. The 4 document representation techniques are: T1 - remove stop words, T2 - tag document as entity types (person, location, and organization), T3 – apply T1 and remove all other words except nouns and T4 – apply T3 and tag terms as entity types (person, location, and organization). The 5 enrichment techniques are: E1 – get top similar articles, add titles and categories, E2 – query Wikitology using Lucene, add titles, categories and linked concepts, E3 – apply E2 and query Freebase on entity types, E4 – Filter results returned by E1, E2 and E3 on defined criteria; and E5 – remove noise and delimiters from results returned by E1, E2, E3 and E4. Comprehensive experiments are carried out on 20 Newsgroup dataset with different combinations of text document representation (T1-T4) and enrichment (E1-E5) techniques. Experiments were evaluated on Support Vector Machine (SVM) using 10 fold cross-validations. In this paper we are representing same experiments evaluation on Naïve Bayes and will compare results of both SVM and NB.

Now, we will like to give brief description of machine learning techniques used by us (SVM and NB), for our experiments evaluation. The Support Vector Machines is a classifier that finds best hyper plane between two classes of data, by separating positive and negative examples through solid line in the middle called decision line. In following figure; gap between solid and dashed line reflects the margin of movement of decision line left or right without miss-classification of document [3]. Naive Bayes classifier is basically a probabilistic classifier based on hypothesis. On the basis of assumption and training document; Bayesian learning is to find most appropriate assumption based on prior hypothesis and initial knowledge. Main assumption is that terms in test document have no relation among them and probability is calculated that document belong to category C [1] [2] [3].

### III. EXPERIMENTAL EVAULTION

In this section we explain the experimental setup, the dataset used and the approach used for this research.

#### A. Data

We use the dataset of 20 News group . It is a well-balanced data set of 20 categories with 1000 documents in each.

#### B. Experimental Results

Baseline for the experiment is setup by removing stop words, delimiters and stemmed the dataset by using Porter Stemmer. In baseline we haven't add any semantic background knowledge [12]. We are picking results evaluated on SVM from [12] and analyzing the same experiments on NB. We have use micro-average and macro-average F measure. For evaluating data set we have use 10 fold cross- validation and used paired t-test to assess the significance.

Table 1 shows the performance of dataset by augmenting knowledge (top 5 and 20 categories and titles) using combined enrichment technique E1, E4 and E5 on text document representation T1. In table 2 Baseline shows the dataset with no semantic background knowledge, A1 means top 5 titles and their related categories, A2 means top 20 titles and their related categories. Table 2 illustrates the performance of dataset by integrating knowledge (top 5 and 20 categories, titles and linked concepts) using combined enrichment technique E2, E4 and E5 on text document representation T1. In table 3 Baseline is same, A3 means top 5 titles, their related categories and linked concepts, A4 means top 20 titles, their related categories and linked concepts. Table 3 demonstrates the performance of dataset by incorporating data (top categories, titles and linked entities) using combined enrichment technique E1, E2, E4 and E5 on text document representation T1. In table 4 Baseline remains the same, A5 means top 20 titles, their related categories and linked concepts. Table 1, 2, 3 in general illustrates the micro-average F-measure, macro-average F- Measure and improvement (or declination) of experiments A1-A5 after applying text classification technique SVM [12] and Naïve Bayes.

The best improvement up to date by adding semantic background knowledge to enhance text categorization is of 0.919 (micro-average F-Measure) / 0.920 (macro-average F-Measure) with improvement of +5.88% and +6.36%, respectively as compare to simple baseline using SVM text classifier [12]. Using Naïve Bayes classifier 0.881 (micro-average F-Measure) / 0.877 (macro-average F-Measure) with improvement of +27.12% and +28.78%, respectively as compare to simple baseline. This improvement is achieved on A4 in which we have add top 20 titles, categories and their linked concepts and also applying filtration criteria defined in enrichment technique E4 and E5[12]. The worst results is achieved on A2 by adding top 20 titles and their related categories using enrichment techniques E1, E4 and E5[12]. From [12] using SVM classifier 0.770 (micro-average F-Measure) / 0.757 (macro-average F-Measure) with declination of -11.29% & -12.49% respectively. Using Naïve Bayes text classifier on 0.685 (micro-average F-Measure) / 0.676 (macro-average F-Measure) with declination of -1.15% & -0.73% respectively. Figure 1 is a graphical representation of micro-average and macro-average F-measure of all approaches (including baseline) comparing micro-average and macro-average F- measure of SVM and Naïve Bayes classifier. Figure 2 shows the percentage of improvement or declination of micro-average and macro-average F-Measure of SVM and Naïve Bayes classifier, after integrating knowledge from Wikitology.

News Group dataset. Improvement from 0.649 to 0.714 and 0.667 to 0.719 is shown on Reuters-21578 and 20 News Group dataset respectively. Improvement in text clustering was also shown by Hotho, Staab, Stumme [9], incorporating knowledge from Word Net. They assimilate background information to text document to reduce the variance in semantics of a document under one category. Evaluation was done on Reuters-21578 using purity and inverse purity measures. Gabrilovich, Markovitch [5] suggested a model in which feature generator is built by mapping terms of document to Open Project Directory (OPD) concepts and features are generated by contextual analysis of the document. Model is evaluated on datasets of Reuters-21578, RCV1, 20 News Group and Movies. Evaluated results shown improvement on all the datasets.

TABLE 1: Percentage improvement over SVM

|  | SVM |  | SVM -Improvement |  |
|---|---|---|---|---|
| **Enrichment** | **Micro** | **Macro** | **Micro** | **Macro** |
| **Baseline** | 0.868 | 0.865 | - | - |
| **A1** | 0.784 | 0.768 | -9.68% | -11.21% |
| **A2** | 0.770 | 0.757 | -11.29% | -12.49% |
| **A3** | 0.843 | 0.830 | -2.88% | -4.05% |
| **A4** | 0.919 | 0.920 | **5.88%** | **6.36%** |
| **A5** | 0.851 | 0.839 | -1.96% | -3.01% |

TABLE 2: Percentage improvement over NB

|  | Naïve Bayes(NB) |  | NB -Improvement |  |
|---|---|---|---|---|
| **Enrichment** | **Micro** | **Macro** | **Micro** | **Macro** |
| **Baseline** | 0.693 | 0.681 | - | - |
| **A1** | 0.687 | 0.676 | -0.86% | -0.73% |
| **A2** | 0.685 | 0.676 | -1.15% | -0.73% |
| **A3** | 0.693 | 0.681 | 18.75% | 18.94% |
| **A4** | 0.823 | 0.820 | **27.12%** | **28.78%** |
| **A5** | 0.804 | 0.792 | 16.01% | 16.29% |

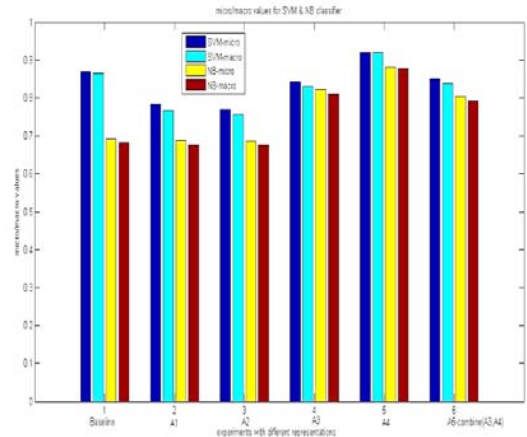

**Figure 1 Graphical Representation of experimental results.**

IV. FUTURE WORK & CONCLUSION

In this paper we present an enhanced version of our previous research [12] by comparing experiment results on SVM and Naïve Bayes Classifier. In [12] we have evaluated our experiments on SVM, in which we found the improvement from 0.868 to 0.919 (micro-average f-measure) and 0.865 to 0.920 (macro-average f-measure). After evaluating same experiments on Naïve Bayes classifier, we found the improvement from 0.693 to 0.881 (micro-average f-measure) and 0.681 to 0.877 (macro-average f-measure). Clear improvement of 6.36% and 28.78% is achieved on SVM and Naïve Bayes classifier respectively, by integrating information extracted from Wikitology.

For future works we can carry out the same experiments on different datasets like Reuters-21578, RCV1, OHSUMED, Movies etc and can also apply other text categorization algorithms like K- Nearest Neighbor, Matrix Regression, and Decision trees etc. In-depth analysis and comparison can be holding on the diverse datasets and text classification techniques.